\documentclass{article}

\usepackage{microtype}
\usepackage{xcolor}
\usepackage{multirow}
\usepackage{mathtools}
\usepackage{amsfonts}
\usepackage{graphicx}
\usepackage{subfigure}
\usepackage{booktabs} 
\usepackage{amsmath}
\usepackage{bbm}
\usepackage[bottom]{footmisc}

\usepackage{hyperref}

\usepackage[accepted]{icml2021}

\icmltitlerunning{Developed under 2023 MC REU}

\usepackage{amsmath,amsfonts,bm}

\def\eqref#1{equation~\ref{#1}}

\def\1{\bm{1}}

\DeclareMathAlphabet{\mathsfit}{\encodingdefault}{\sfdefault}{m}{sl}
\SetMathAlphabet{\mathsfit}{bold}{\encodingdefault}{\sfdefault}{bx}{n}

\newcommand{\softmax}{\mathrm{softmax}}

\DeclareMathOperator*{\argmax}{arg\,max}
\DeclareMathOperator*{\argmin}{arg\,min}

\usepackage{amsthm}
\theoremstyle{plain}

\theoremstyle{definition}

\begin{document}

\twocolumn[
\icmltitle{CS-Mixer: A Cross-Scale Vision MLP Model with Spatial–Channel Mixing}



\icmlsetsymbol{equal}{*}

\begin{icmlauthorlist}


\icmlauthor{Jonathan Cui}{psu}
\icmlauthor{David A. Araujo}{psu}
\icmlauthor{Suman Saha}{psu}
\icmlauthor{Md. Faisal Kabir}{psu}
\end{icmlauthorlist}

\icmlaffiliation{psu}{The Pennsylvania State University.}

\icmlcorrespondingauthor{Md. Faisal Kabir}{mpk5904@psu.edu}

\icmlkeywords{Machine Learning, ICML}

\vskip 0.4in
]



\printAffiliationsAndNotice{}  

\begin{abstract}
Despite their simpler architectural designs compared with Vision Transformers and Convolutional Neural Networks, Vision MLPs have demonstrated strong performance and high data efficiency in recent research. However, existing works such as CycleMLP and Vision Permutator typically model spatial information in equal-size spatial regions and do not consider cross-scale spatial interactions. Further, their token mixers only model 1- or 2-axis correlations, avoiding 3-axis spatial\textendash channel mixing due to its computational demands. We therefore propose CS-Mixer, a hierarchical Vision MLP that learns dynamic low-rank transformations for spatial\textendash channel mixing through cross-scale local and global aggregation. The proposed methodology achieves competitive results on popular image recognition benchmarks without incurring substantially more compute. Our largest model, CS-Mixer-L, reaches 83.2\% top-1 accuracy on ImageNet-1k with 13.7 GFLOPs and 94 M parameters.
\end{abstract}

\begin{figure}[t]
    \centering
    \includegraphics[width=\columnwidth]{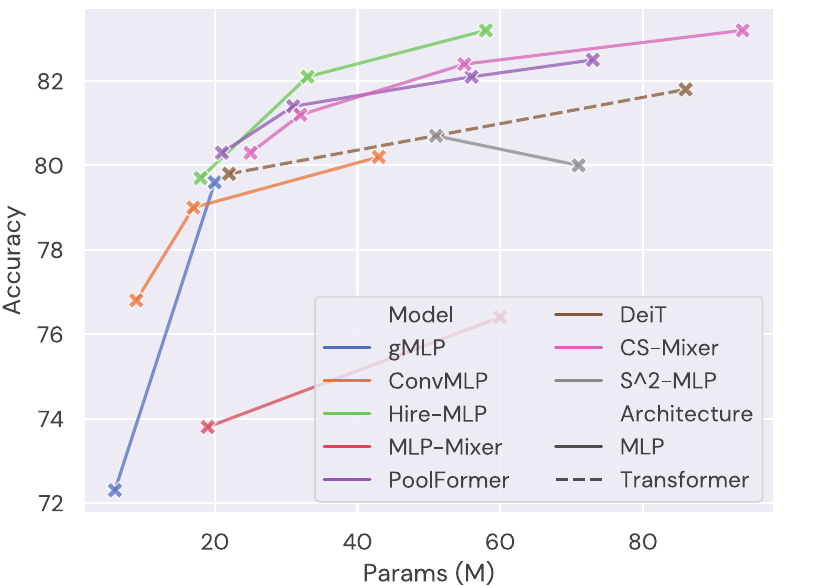}
    \vskip -0.08in
    \caption{ImageNet-1k top-1 accuracy vs.\ model size, with resolution $224\times224$ and no extra training data.}
    \label{fig:acc-vs-param}
\end{figure}

\begin{figure*}[t]
  \centering
  \includegraphics[width=0.9\textwidth]{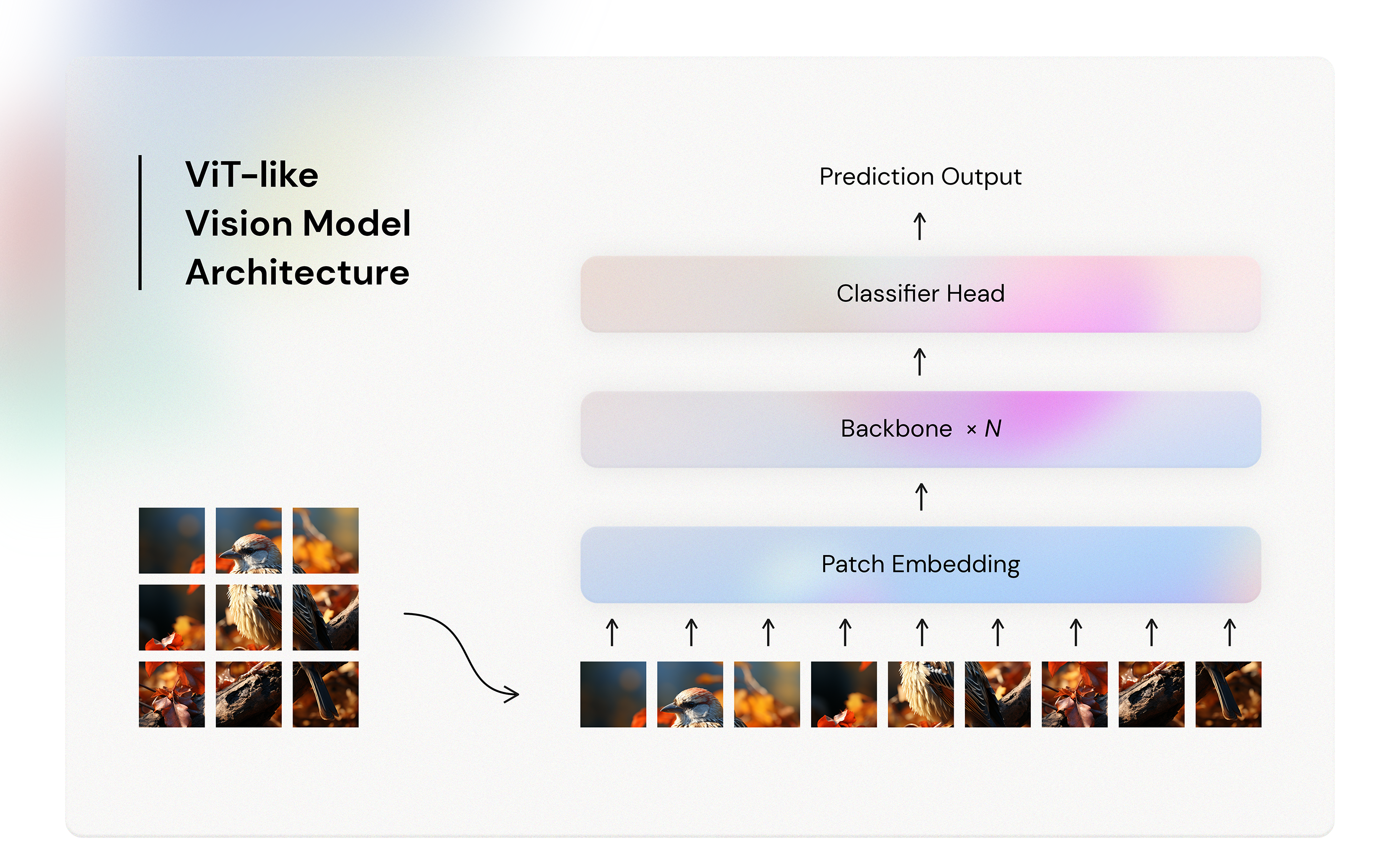}
  \vskip -0.08in
  \caption{The general architecture of Vision Transformers \citep{dosovitskiy2021an}.}
  \label{fig:vit-arch}
\end{figure*}

\section{Introduction}

The field of Computer Vision seeks to model the mechanisms by which humans perceive and interpret visual signals. To this end, researchers devise models, typically neural networks, that learn from a large dataset of images accompanied with labels, annotations, etc. and measure their performance on various benchmarks. For a long time, \hyphenation{Convo-lu-tion-al} neural networks (CNNs), which use discrete convolutions and pooling operations to parameterize networks with spatial inductive biases, have seen extensive applications since they were popularized by AlexNet \citep{krizhevsky2012alexnet}. The introduction of such spatial locality and translational invariance to the model has seen immense success in earlier deep learning literature, and many CNNs, e.g., EfficientNets \citep{tan2019efficientnet}, NFNets \citep{brock2021nfnet}, and ConvNeXts \citep{liu2022convnext}, remain crucially relevant in Computer Vision today.

At the beginning of 2020, \citet{dosovitskiy2021an} revolutionized the field with Vision Transformers (ViTs), borrowing directly from Natural Language Processing (NLP) paradigms to construct neural networks. Specifically, they view an input image as a sequence of smaller (e.g., $7\times7$) non-overlapping patches, where each patch is treated as a token. The resulting representations are thus a sequence of tokens, the space on which NLP models operate. Then, the tokens are passed through a Transformer encoder and finally sent to task-dependent heads for downstream applications. As ViTs have shown, vision neural networks need not rely on handcrafted spatial equivariances to achieve robust performance; rather, sufficiently expressive spatial operations like self-attention \citep{vaswani2017attention} can replace the role of convolution and pooling without any performance degradation. With these advantages, ViTs set a new paradigm for vision models with its structure (Fig.~\ref{fig:vit-arch}): a patch embedding layer followed by a number of backbone layers, each comprising a token mixer and a channel mixer, and finally a task-dependent head for, e.g., classification. Many other Vision Transformer architectures, such as Swin Transformers \citep{liu2021swin}, CaiT \citep{touvron2021cait}, and MaxViT \citep{tu2022maxvit}, followed this line of work and achieved stronger performance in various vision benchmarks.

Another line of work, led by MLP-Mixers \citep{tolstikhin2021mlp}, explored the use of similar vision architectures with only Multi-Layer Perceptrons (MLPs). MLPs are the origin of, and remain ubiquitous in, all neural network architectures, and in particular they serve as ViTs' channel mixers. Recognizing that convolution-like structures are unnecessary for vision models, MLP-Mixers proposed to replace the self-attention mechanism in ViTs, resulting in a fully-MLP architecture. In particular, MLP-Mixers use MLPs for its token mixers, which model spatial transformations in the same way channel information is processed. However, the simple token-mixing design, which does not capture any high-order interaction, proved inadequate to modeling complex cross-token information.

A large body of subsequent works \citep{chen2022cyclemlp,guo2022hiremlp,lian2022asmlp,hou2022vip} focused on designing MLP token mixers that incorporate spatial prior information to boost performance. These MLP-based models mainly focus on token mixing within static regions of fixed sizes to learn progressively richer information. Despite competitive performance with state-of-the-art Transformer architectures, fixed-size spatial mixing mechanisms cannot attend to objects of various sizes, nor does their general architecture model explicit cross-layer information to obtain cross-scale image representations.

Further, modern Vision MLP models typically learn token-mixing transformations only over one or two axes of rank-3 image representations: channel-only \citep{chen2022cyclemlp,yu2022s2mlp}, height and width \citep{tolstikhin2021mlp,touvron2021resmlp}, or a combination of the channel axis and either spatial axis \citep{hou2022vip,guo2022hiremlp}. This is largely due to a limited computational budget, as mixing height, width, and channels simultaneously results in at least $O(N^4D^2)$ number of parameters for fully-connected layers ($N=\text{resolution}$, $D=\text{embed. dim}$). While some latest architectures are exceptions \citep{go2023gswin,zheng2022msmlp}, they still fail to model cross-scale information at large. We therefore raise the question: can we scale MLP-based vision neural networks further to model 3-axes spatial\textendash channel interactions without degrading network performance or incurring additional computational cost?

To this end, we introduce a new MLP-based vision architecture dubbed CS-Mixer, which leverages multi-scale spatial information and directly parametrizes expressive non-linear transforms across all three axes to model spatial token mixing. Specifically, by incorporating cross-scale information through local and global aggregation, CS-Mixers managed to learn expressive transformations over height, width, and channels simultaneously with limited compute. The resulting models are competitive with state-of-the-art Transformer- and MLP-based models, as shown in Fig.~\ref{fig:acc-vs-param}, without incurring much more parameters or computational FLOPs. We expect this work to set a benchmark for future works and guide further efforts in investigating 3-axis token mixing strategies.

\section{Related Works}

\paragraph{Vision Transformers.} Motivated by the success of Transformer architectures in NLP, \citet{dosovitskiy2021an} proposed to introduce Transformers to computer vision datasets, leveraging the power of self-attention to learn rich data representations. Following this work, CaiTs \citep{touvron2021cait} and DeepViTs \citep{zhou2021deepvit} resolved the saturation of performance when scaling the network depth, which showed promising results for the scalability of vision models similar to that discovered in NLP paradigms. Later literature expanded on this idea and adopted pyramid architectures as commonly used in CNNs \citep{liu2021swin,pan2021hvt,xu2021vitae}, reducing the number of tokens and introducing vision-specific inductive biases like multi-granularity to enhance model performance. Unlike Vision MLPs, which use only MLP-based operations, Vision Transformers always rely on the Self-Attention operation \citep{vaswani2017attention} to achieve token mixing.

\paragraph{Vision MLP Architectures.} As Transformer-based architectures gained increasing traction in modern Computer Vision \citep{dosovitskiy2021an}, a number of works began to investigate different token-mixing strategies, replacing the computationally heavy $O(N^4)$ multi-head self-attention layers with purely MLP-based operations \citep{tolstikhin2021mlp,liu2021gmlp,touvron2021resmlp}. In parallel with Vision Transformers, the development of Vision MLPs also focused on spatial inductive biases. For instance, CycleMLPs \citep{chen2022cyclemlp} and ActiveMLPs \citep{wei2022activemlp} used deformable convolutions to introduce locality biases,\footnote{The deformable convolution used is a degenerate form that is equivalent to a general formulation of MLPs.} and Vision Permutators \citep{hou2022vip} and Hire-MLPs \citep{guo2022hiremlp} achieved local token mixing with tensor reshaping. However, these methods are almost always limited to 1- or 2-axis token mixing, since the number of parameters, which is quadratic in the size of the axis, is not readily scalable. Therefore, there remains a lack of exploration in the field in terms of 3-axis, spatial\textendash channel token-mixing strategies.



\section{Methodology}

\paragraph{Notation.} We denote tensors with lowercase letters in italics, e.g., $x,y,z$; matrices are sometimes denoted with capital, italicized letters. We use the notation $[n]$ to represent the set $\{1,\cdots,n\}$ for a positive integer $n$.

\subsection{Preliminaries}

The task of image classification is formulated as follows. Suppose a finite, labeled $K$-class training dataset $\mathcal D\subset\mathcal X\times\mathcal Y$ is given, where $\mathcal X=\mathbb R^{H\times W\times C}$ is the input image space and $\mathcal Y=[K]$ is the output space. A neural network architecture is a collection of parametrized maps
\begin{equation*}
  \{F_\theta\colon\mathcal X\to\tilde{\mathcal Y}\mid \theta\in\Theta\},
\end{equation*}
where $\tilde{\mathcal Y}=\mathbb R^K$ is the logit space, and $\Theta$ is the parameter space defined by the neural network. By applying the softmax operation, we obtain a probability distribution $p_\theta(y\mid x)$ over the possible classes for the dataset:
\begin{equation}
  p_\theta(y\mid x)\coloneqq\softmax(F_\theta(x))_y,\quad~~\forall(x,y)\in\mathcal X\times\mathcal Y.
\end{equation}
We then minimize the cross-entropy loss function over the training dataset via the AdamW first-order optimizer \citep{loshchilov2017adamw}:
\begin{align}
  &\theta^*\coloneqq\argmin_{\theta\in\Theta}\mathcal L(\theta),\text{~where}\\
  &\mathcal L(\theta)\coloneqq-\frac1{\lvert{\mathcal D}\rvert}\sum_{(x^{(i)},y^{(i)})\in\mathcal D}H[p_{y^{(i)}}(y),p_\theta(y\mid x^{(i)})],\label{eq3}
\end{align}
where $H[\cdot,\cdot]$ is the cross entropy and $p_{y^{(i)}}(y)$ is a $\delta$ distribution centered at $y^{(i)}$.

To measure the performance of a trained neural network $F_\theta$, we take a testing dataset $\mathcal D'\subset\mathcal X\times\mathcal Y$ and report the accuracy:
\begin{equation}
  \text{Acc}(\theta)\coloneqq\frac{100}{\lvert{\mathcal D'}\rvert}\sum_{(x^{(i)},y^{(i)})\in\mathcal D'}\mathbb I[y^{(i)}=\argmax_kp_\theta(k\mid x^{(i)})],\label{eq:acc}
\end{equation}
where $\mathbb I[\cdot]$ is the Iverson bracket.

\begin{figure}[t]
  \centering
  \includegraphics[width=\columnwidth]{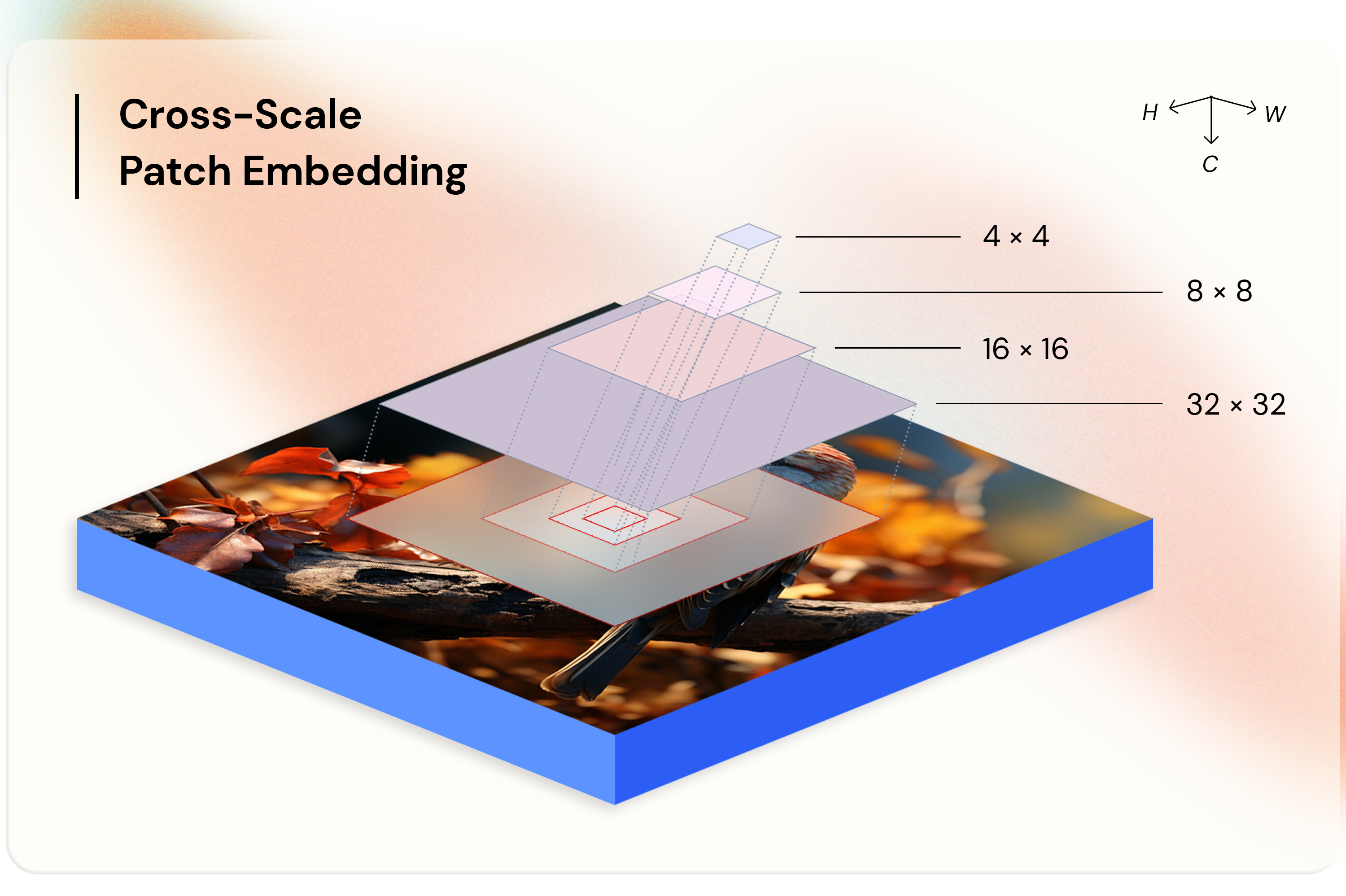}
  \vskip -0.08in
  \caption{An illustration of CS-Mixer's cross-scale patch embedding mechanism.}
  \label{fig:ce-patch-embed}
\end{figure}

\subsection{Network Architecture}

We adopt a common hierarchical Vision MLP structure as in Fig.~\ref{fig:vit-arch}, with some modifications. To leverage cross-scale spatial information, we follow \citet{wang2022crossformer} and apply a cross-scale embedding layer to an $H\times W\times C$ input image with patch sizes $4,8,16,32$, yielding image embeddings of size $H/4\times W/4$. The tokens, viewed as an $h\times w\times c$ tensor which represents the image, are then passed through 4 backbone stages, interleaved with cross-scale patch merging layers to aggregate and communicate cross-scale information for spatial downsampling. A backbone stage comprises a varying number of stacked CS-Mixer layers, each involving a token-mixer and a channel-mixer. We use the proposed CS-Mixer operator for token mixing, and set the channel-mixer to a three-layer MLP with MLP ratio 4 following the convention by ViTs.

\paragraph{Cross-Scale Embedding Layer.} We adopt a cross-scale embedding layer similar to \citet{wang2022crossformer} and \citet{tatsunami2022raftmlp} with patch sizes 4, 8, 16, and 32, as shown in Fig.~\ref{fig:ce-patch-embed}. In particular, given an input image of size $H\times W\times C$, we apply 4 convolutional layers with stride 4 and kernel sizes 4, 8, 16, and 32, where padding is applied such that the outputs share spatial dimensions $(H/4)\times(W/4)$. For embedding dimension $c$, we set the number of output channels to $c/2$, $c/4$, $c/8$, and $c/8$ respectively, and then concatenate along the channel axis to produce embeddings of size $(H/4)\times(W/4)\times c$. Observe that the order of concatenation does not matter as any permutation may be readily absorbed by subsequent layers.

\paragraph{Backbone Stages.} The backbone of the CS-Mixer has a hierarchical pyramid structure as adopted by \citet{liu2021swin}. The $i$-th backbone stage $(i\in[4]$) is the composition of $L_i$ identical but separately parametrized CS-Mixer layers, where each layer is a stack of a token mixer and a channel mixer with residual connections. Note that backbone stages do not modify the shape of input embeddings.

\paragraph{Cross-Scale Patch Merging Layer.} We use a cross-scale embedding layer with patch sizes 2 and 4 instead for patch merging, where the stride is set to the minimum patch size, i.e., 2. Input embeddings of shape $h\times w\times c$ have output shape $(h/2)\times(w/2)\times(2c)$, thus reducing each spatial dimension by a factor of 2 and doubling the number of channels for the next stage.

\paragraph{Classifier Head.} We adopt a standard classifier head which performs global averaging and then projects linearly to $\mathbb R^K$ to produce $K$-class logits.

\subsection{CS-Mixer Operator}

Given a group size $g$ which divides $h$ and $w$, the CS-Mixer operator takes input embeddings $x\in\mathbb R^{h\times w\times c}$ and rearranges $x$ into $N=hw/g^2$ groups of $L\times c$ sequence embeddings ($L=g^2$) by aggregating local or global tokens. This rearrangement may be described in $\tt{einops}$ notation \citep{rogozhnikov2022einops} as follows:
\begin{align*}
  \text{LA:}\quad\tt{(n_h~g_1)~(n_w~g_2)~c~\to~(n_h~n_w)~(g_1~g_2)~c},\\
  \text{GA:}\quad\tt{(g_1~n_h)~(g_2~n_w)~c~\to~(n_h~n_w)~(g_1~g_2)~c},
\end{align*}
where LA and GA refer to local and global token aggregation respectively. This is similar to the concurrent work by \citep{wang2022crossformer}.

The groups of aggregated tokens $\{\tilde x^k\in\mathbb R^{L\times c}\}_{k\in[N]}$ are then mixed in parallel with $m$ heads. For each sequence of tokens $\tilde x^k\in\mathbb R^{L\times c}$, we calculate:
\begin{align}
  &u\gets\tilde x^k\cdot W_u,~~v^n\gets\tilde x^k\cdot W^n_v,\phantom\sum\label{eq2.1}\\[.45em]
  &v^n_{k,l}\gets\sum_{i,j}v^n_{i,j}\cdot w^n_{i,j,k,l},\label{eq2.2}\\
  &v\gets\sum_{n}v^n\cdot\tilde W^n_v,~~y\gets(u*v)\cdot W_o\label{eq2.3},
\end{align}
where $W_u,W_o\in\mathbb R^{c\times c}$, $W^n_v\in\mathbb R^{c\times d}$, $\tilde W^n_v\in\mathbb R^{d\times c}$, and $w^n\in\mathbb R^{L\times d\times L \times d}$ are weight matrices that parametrize the CS-Mixer operator.\footnote{We omit biases and norm layers for brevity.} 


\begin{table}[t]
  \centering
  \vskip -0.5em
  \caption{Different configurations of CS-Mixer: Tiny, Small, Base, and Large.}
  \label{table:model-sizes}
  \vskip 1em
  \begin{small}\begin{tabular}{l|cccc}
    \toprule
    & \textbf{T}iny & \textbf{S}mall & \textbf{B}ase & \textbf{L}arge \\
    \midrule
    Param (M) & 25.4 & 32.2 & 55.9 & 94.2 \\
    GFLOPs & 2.4 & 4.2 & 7.8 & 13.7 \\
    \midrule
    Base dim. ($C$) & 64 & 96 & 96 & 128 \\
    Rank ($d$) & 2 & 4 & 4 & 4 \\
    \#Layers ($L_i$) & 1,1,8,6 & 2,2,6,2 & 2,2,18,2 & 2,2,18,2 \\
    \#Heads ($m$) & 2,4,8,16 & 3,6,12,24 & 3,6,12,24 & 4,8,16,32 \\
    \bottomrule
  \end{tabular}\end{small}
\end{table}

\renewcommand*{\thefootnote}{\fnsymbol{footnote}}
\setcounter{footnote}{0}
\begin{table*}[t!]
  \centering
  \begin{small}\begin{tabular}{l|cccc}
    \toprule
    Model & Mixing & Param (M) & GFLOPs & Accuracy \\
    \midrule
    MLP-Mixer-S/16 & \multirow{2}{*}{\textit{H}\textendash \textit{W}} & 19 & 3.8 & 73.8 \\
    MLP-Mixer-B/16 & & 60 & 11.7 & 76.4 \\
    \midrule
    gMLP-T & \multirow{3}{*}{\textit{H}\textendash \textit{W}} & 6 & 1.4 & 72.3 \\
    gMLP-S & & 20 & 4.5 & 79.6 \\
    gMLP-B & & 73 & 15.8 & 81.6 \\
    \midrule
    ConvMLP-S & \multirow{3}{*}{N/A} & 9 & 4.8 & 76.8 \\
    ConvMLP-M & & 17 & 7.8 & 79.0 \\
    ConvMLP-L & & 43 & 19.8 & 80.2 \\
    \midrule
    Hire-MLP-T & \multirow{3}{*}{(\textit{H}/\textit{W})\textendash \textit{C}} & 18 & 2.1 & 79.7 \\
    Hire-MLP-S & & 33 & 4.2 & 82.1 \\
    Hire-MLP-B & & 58 & 8.1 & 83.2 \\
    \midrule
    PoolFormer-S24 & \multirow{4}{*}{\textit{H}\textendash \textit{W}\footnotemark} & 21 & 3.6 & 80.3 \\
    PoolFormer-S36 & & 31 & 5.2 & 81.4 \\
    PoolFormer-M36 & & 56 & 9.1 & 82.1 \\
    PoolFormer-M48 & & 73 & 11.9 & 82.5 \\
    \midrule
    $\text{S}^{\text{2}}$-MLP-W & \multirow{2}{*}{\textit{C}} & 71 & 14.0 & 80.0 \\
    $\text{S}^{\text{2}}$-MLP-D & & 51 & 10.5 & 80.7 \\
    \midrule
    CS-Mixer-T & \multirow{4}{*}{\textit{H}\textendash\textit{W}\textendash\textit{C}} & 25 & 2.4 & 80.3 \\
    CS-Mixer-S & & 32 & 4.2 & 81.2 \\
    CS-Mixer-B & & 56 & 7.8 & 82.4 \\
    CS-Mixer-L & & 94 & 13.7 & 83.2 \\
    \bottomrule
  \end{tabular}\end{small}
  \caption{Comparison of state-of-the-art vision models on ImageNet-1k \citep{russakovsky2015imagenet} at resolution $224\times224$ without extra data.}
  \label{table:imagenet-1k}
\end{table*}

The mixing achieved by elementwise multiplication is inspired by gMLPs \citep{liu2021gmlp}. However, a key distinction is that we perform the spatial\textendash channel mixing of $v$ in a low-dimensional subspace $\mathbb R^L\otimes\mathbb R^d$. Noting that $w^n\in\mathbb R^{L\times d\times L\times d}\cong\hom(\mathbb R^L\otimes\mathbb R^d,\mathbb R^L\otimes\mathbb R^d)$ in Equation~\ref{eq2.2}, where the subspace $\mathbb R^L\otimes\mathbb R^d$ is obtained by the image under $W^n_v$. Simply including all axes in the mixing process would result in five to six orders of magnitude more parameters than channel-only mixing strategies. Therefore, to reduce the number of parameters while retaining expressivity, we directly parameterize low-rank transforms on $\mathbb R^L\otimes\mathbb R^c$ by first projecting $\mathbb R^c$ to a lower-dimensional space $\mathbb R^d$ ($d\ll c$), where $d$ controls the rank of the transform. Our experiments demonstrate that dimensionalities as low as $d=2$ or $d=4$ suffice to capture the complexities arising in the task of image recognition.

\section{Experiments}

\begin{table}[t]
  \centering
  \begin{small}\begin{tabular}{l|cc}
    \toprule
    Model & CIFAR-10 & CIFAR-100 \\
    \midrule
    Mixer-B/16 & 97.7 & 85.0 \\
    \midrule
    RaftMLP-S & 97.4 & 85.1 \\
    RaftMLP-M & 97.7 & 86.8 \\
    RaftMLP-S & 98.1 & 86.8 \\
    \midrule
    CS-Mixer-T & 98.36 & 87.86 \\
    CS-Mixer-S & 98.21 & 88.03 \\
    CS-Mixer-B & 98.59 & 89.41 \\
    CS-Mixer-L & 98.63 & 89.73 \\
    \bottomrule
  \end{tabular}\end{small}
  \caption{Transfer learning results on the CIFAR-10 and CIFAR-100 datasets at resolution $224\times224$, pretrained on ImageNet-1k.}
  \label{table:cifar}
\end{table}

\subsection{ImageNet-1k Classification}\label{subsec:imagenet-1k-clsf}

\paragraph{Settings.} We design four models of the CS-Mixer architecture with different sizes, reported in Table~\ref{table:model-sizes}: \textbf{T}iny, \textbf{S}mall, \textbf{B}ase, and \textbf{L}arge. The models are implemented in PyTorch \citep{paszke2017pytorch} and trained on the ImageNet-1k dataset \citep{russakovsky2015imagenet} without extra data at resolution $224\times 224$. We employ a standard training procedure of 300 epochs and adopt RandAugment \citep{cubuk2020randaug}, Mixup \citep{zhang2018mixup}, CutMix \citep{yun2019cutmix}, Random Erasing \citep{zhong2020randomerase}, and Stochastic Depth \citep{huang2016stochdepth} for augmentation and regularization. Note that Mixup and CutMix change the $p_{y^{(i)}}(y)$ term in the loss function (Equation~\ref{eq3}). We perform $L_2$ regularization at rate $5\times10^{-2}$, and adopt 20 warmup epochs with learning rate $10^{-6}$ and 10 cooldown epochs with learning rate $10^{-5}$. We use the cosine learning rate scheduler \citep{loshchilov2016cosine} stepped every epoch with global batch size 1024, base learning rate $2\times10^{-3}$, and minimum learning rate $10^{-5}$. We also use the exponential moving average weights for validation \citep{cai2021ema}, with decay rate $\gamma=.99996$. We report the accuracy (Equation~\ref{eq:acc}) on the validation set of 50,000 images with a 90\% center cropping for image preprocessing.

\footnotetext{PoolFormers' token mixer transforms both spatial axes but is not parametrized. We therefore view it as a trivial Vision MLP.}

\subsection{Transfer Learning}

We also apply transfer learning on the CIFAR-10 and CIFAR-100 datasets \cite{cifar10}. We redefine the model's classifier head with the correct number of classes, which is randomly initialized. The models are then trained with the same recipe as in Subsection~\ref{subsec:imagenet-1k-clsf} with the following hyperparameters replaced: global batch size $256$, learning rate $10^{-4}$, weight decay $10^{-4}$, and crop percentage 100\%. In particular, we resize all images to $224\times224$ during training and evaluation with no cropping. We also disable exponential moving average for transfer learning experiments, and compare with results reported by \citet{tatsunami2022raftmlp}, which adopted a comparable setup.

\subsection{Discussion}

\begin{figure}[b!]
    \centering
    \includegraphics[width=\columnwidth]{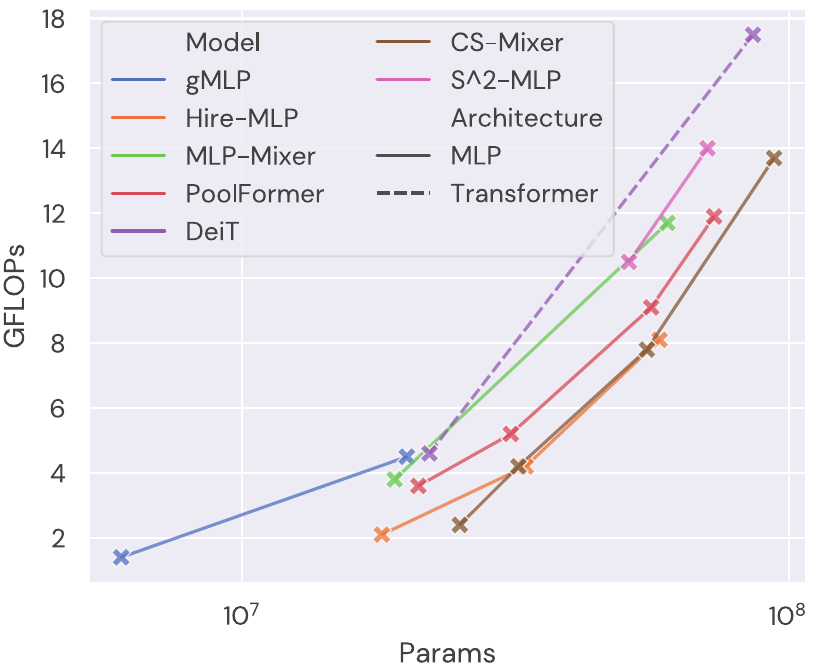}
    \vskip -0.08in
    \caption{Model FLOP count vs.\ number of parameters. CS-Mixers require competitively low compute at the same number of parameters compared with previous models.}
    \label{fig:flops-vs-params}
\end{figure}

As demonstrated in Table~\ref{table:imagenet-1k}, CS-Mixers across sizes achieve robust performance on ImageNet-1k when compared against state-of-the-art vision models. While 3-axis mixing incurs more parameters, our models maintain comparable FLOP count with models that are much smaller by modeling spatial\textendash channel mixing in a low-rank subspace (Equation~\ref{eq2.2}). For instance, CS-Mixer-L is larger than $\text{S}^{\text{2}}$-MLP-W by over 20 million parameters, while requiring marginally lower compute measured by FLOPs (Figure~\ref{fig:flops-vs-params}). As we continue this work, we expect to further investigate the effect of the rank on network performance as well as its theoretical implications.

\begin{figure}[t]
    \centering
    \includegraphics[width=\columnwidth]{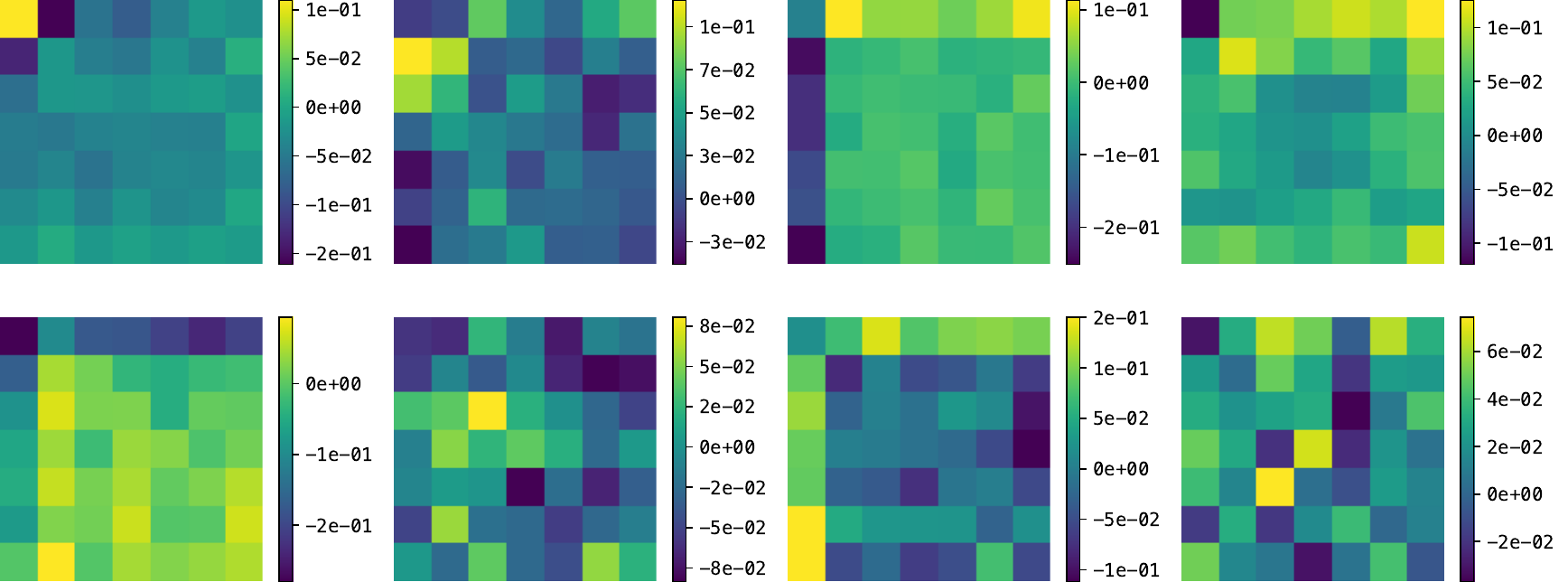}
    \vskip -0.08in
    \caption{A visualization of how the first output neuron connects to inputs from the first channel in the first head, from the four stages (columns). The first row comes from the first LA layer in each stage and the second from the first GA layer respectively.}
    \label{fig:weights}
\end{figure}

We also visualize the spatial weights $w^{1}_{:,1,1,1}\in\mathbb R^L\cong\mathbb R^{g\times g}$, without loss of generality, learned by our token mixers for LA and GA in Figure~\ref{fig:weights}. The weights exhibit non-trivial patterns across both local and global scales along spatial axes, which corroborates the effectiveness of local and global aggregation for token mixing. 



\section{Conclusions}

This paper proposes a novel Vision MLP architecture dubbed CS-Mixer, which adopts cross-scale token aggregation to achieve spatial\textendash channel token-mixing for an expressive vision backbone. The incorporation of cross-scale patch embedding and patch merging layers allows our models to learn meaningful spatial interactions across scales, and local/global aggregations make it possible to model 3-axis mixing without incurring intractable computational costs. We recognize much space for future investigation, such as ablation studies, learned internal representations, and training dynamics, and look forward to expanding this work in the coming months. We expect this work to set a benchmark for future works in image recognition and guide further efforts in studying effective token mixing strategies.

\section*{Acknowledgements}

We would like to express our sincere gratitude for Penn State University's \textit{Multi-Campus Research Experience for Undergraduates} program and Penn State Harrisburg's \textit{Pathway to Success: Summer Start} program for the funding that made this research possible. We are deeply indebted to Dr. Kabir and Dr. Saha for their guidance on this work, and we also thank Dr. McHugh for his help with the mathematical aspects related to this paper.



\bibliography{main}
\bibliographystyle{icml2021}





\end{document}